\documentclass[conference]{IEEEtran}
\usepackage[top=1in, left=0.75in, right=0.75in, bottom=1in]{geometry}
\usepackage{etoolbox}
\usepackage{graphicx}
\usepackage{subfigure}
\usepackage{fancyhdr}

\usepackage{float}
\usepackage{amsmath}
\usepackage{multirow}
\usepackage{colortbl}
\usepackage[table]{xcolor}
\usepackage{fontawesome}

\usepackage{algorithm}
\usepackage{algorithmic}
\usepackage{overpic}
\usepackage[utf8]{inputenc} 
\usepackage[T1]{fontenc}    
\usepackage{hyperref}       
\usepackage{url}            
\usepackage{booktabs}       
\usepackage{amsfonts}       
\usepackage{nicefrac}       
\usepackage{microtype}      
\usepackage{xcolor}         
\usepackage{amsmath}
\usepackage{amssymb}
\usepackage{enumitem}
\usepackage{makecell}
\usepackage{tikz}  
\usepackage[font=small]{caption}
\definecolor{mygray}{gray}{0.8} 
\usepackage{etoolbox}
\usepackage{titlesec}
\titlespacing*{\title}{0pt}{0pt}{0pt} 

\renewcommand{\IEEEtitletopspaceextra}{0pt}

\patchcmd{\IEEEauthorblockN}{\vskip 0.5em}{\vskip 0.1em}{}{}
\patchcmd{\IEEEauthorblockA}{\vskip 0.5em}{\vskip 0.1em}{}{}

\ifCLASSINFOpdf
\else
\fi
\makeatletter
\newcommand{\linebreakand}{%
  \end{@IEEEauthorhalign}
  \hfill\mbox{}\par
  \mbox{}\hfill\begin{@IEEEauthorhalign}
}
\makeatother

\begin{document}
\renewcommand{\IEEEtitletopspaceextra}{0.375in}

\title{Learning Diffusion Model from Noisy Measurement using Principled Expectation-Maximization Method}

\author{\IEEEauthorblockN{Weimin Bai}
\IEEEauthorblockA{Academy for Advanced \\ Studies,
Peking University\\
weiminbai@stu.pku.edu.cn}
\and
\IEEEauthorblockN{Weiheng Tang}
\IEEEauthorblockA{School of Physics\\
Peking University\\
wayne-tang@stu.pku.edu.cn}
\and
\IEEEauthorblockN{Enze Ye}
\IEEEauthorblockA{College of Future Technology\\
Peking University\\
yez23@stu.pku.edu.cn}
\and
\IEEEauthorblockN{Siyi Chen}
\IEEEauthorblockA{School of Physics\\
Peking University\\
siyichen@stu.pku.edu.cn}
\linebreakand 
\IEEEauthorblockN{Wenzheng Chen}
\IEEEauthorblockA{Wangxuan Institute of Computer \\Technology,
Peking University\\
wenzhengchen@pku.edu.cn}
\and
\IEEEauthorblockN{He Sun}
\IEEEauthorblockA{College of Future Technology\\
Peking University\\
hesun@pku.edu.cn}
\vspace*{-1cm}
}

\maketitle

\begin{abstract}
Diffusion models have demonstrated exceptional ability in modeling complex image distributions, making them versatile plug-and-play priors for solving imaging inverse problems. However, their reliance on large-scale clean datasets for training limits their applicability in scenarios where acquiring clean data is costly or impractical. Recent approaches have attempted to learn diffusion models directly from corrupted measurements, but these methods either lack theoretical convergence guarantees or are restricted to specific types of data corruption. In this paper, we propose a principled expectation-maximization (EM) framework that iteratively learns diffusion models from noisy data with arbitrary corruption types. Our framework employs a plug-and-play Monte Carlo method to accurately estimate clean images from noisy measurements, followed by training the diffusion model using the reconstructed images. This process alternates between estimation and training until convergence. We evaluate the performance of our method across various imaging tasks, including inpainting, denoising, and deblurring. Experimental results demonstrate that our approach enables the learning of high-fidelity diffusion priors from noisy data, significantly enhancing reconstruction quality in imaging inverse problems.

\end{abstract}



\vspace{-0.1in}

\IEEEpeerreviewmaketitle

\section{Introduction}
Diffusion models (DMs)~\cite{ho2020denoising,song2020score,sohl2015deep} are emerging as versatile tools for capturing high-dimensional distributions~\cite{nichol2021improved, ramesh2021zero, ramesh2022hierarchical, rombach2022high, saharia2022palette} and serving as powerful image priors for solving inverse problems~\cite{feng2023score, zhang2023towards, chung2022diffusion, chung2022improving, song2021solving, kawar2023gsure}.
However, training these models typically requires large datasets of clean, high-quality images, which are often expensive or impractical to obtain~\cite{daras2023ambient}. For example, in fluorescent microscopy, low Signal-to-Noise Ratio (SNR) images $\boldsymbol{y}$ are prevalent due to observational corruptions and measurement noise (e.g., motion blur, camera readout noise), while high SNR images $\boldsymbol{x}$ are scarce because of the long exposure times required and the risk of phototoxicity.

Recent works have sought to address this challenge by learning clean DMs directly from noisy measurements. 
For example, RenderDiffusion~\cite{anciukevivcius2023renderdiffusion} and DWFM~\cite{tewari2023diffusion} incorporate the imaging forward model, $f(\cdot):\boldsymbol{x}\to\boldsymbol{y}$, into the generative process to train 3D DMs from 2D images. However, these methods assume multiple measurements per object, limiting their applicability in many real-world scenarios. SURE-Score~\cite{aali2023solving} uses Stein's unbiased risk estimate (SURE) as a regularizer for learning DMs from noisy data. However, this approach heavily constrains the model's parameter space, leading to suboptimal performance when the data are severely corrupted, such as in inpainting tasks. Ambient Diffusion~\cite{daras2023ambient} introduces a further corruption strategy that masks additional pixels to help the DM learn to reconstruct missing information. While effective for inpainting, it struggles with measurement noise and other types of image corruption.

To address these limitations, EMDiffusion~\cite{bai2024expectation} proposes a general expectation-maximization (EM) framework, achieving state-of-the-art performance in learning clean DMs from various types of corrupted images. This framework alternates between an expectation step (E-step), which uses a known diffusion prior to sample the posterior of clean images $\boldsymbol{x}$ given noisy measurements $\boldsymbol{y}$, and a maximization step (M-step), where the recovered clean images are used to refine the diffusion model. However, its E-step relies on an approximate diffusion posterior sampling (DPS) algorithm~\cite{chung2022diffusion}, which lacks theoretical convergence guarantees and sometimes produces low-quality reconstructions, limiting its ability to capture accurate clean image distributions.

In this work, we propose a principled EM framework for learning clean DMs from noisy measurements with arbitrary corruption types. We enhance the E-step of EMDiffusion by upgrading the DPS algorithm to a plug-and-play Monte Carlo (PMC) method, which offers provable convergence guarantees and improved posterior sampling accuracy. This, in turn, leads to more precise learning of clean generative models. Extensive experiments on noisy CIFAR-10 and CelebA datasets with various corruption types demonstrate the effectiveness of our approach.

\section{Methodology}

\subsection{Preliminaries}
Diffusion models (DMs)~\cite{song2020score, ho2020denoising} capture data distributions by approximating the score function, i.e., the gradient of the log-likelihood $\nabla_{\boldsymbol{x}} \log p(\boldsymbol{x})$, using a deep neural network $\boldsymbol{s}_\theta$ parameterized by $\theta$. To achieve this, DMs define a forward stochastic differential equation (SDE) that progressively injects noise, while its solution is a reverse-time SDE that progressively removes the noise:
\vspace{-0.05in}
\begin{equation}
\begin{split}
	&\text{Forward: } \mathrm{d} \boldsymbol{x}=-\frac{\beta_{t}}{2} \boldsymbol{x} \mathrm{d} t+\sqrt{\beta_{t}} \mathrm{~d} \boldsymbol{w} , \\
  &\text{Reverse: } \mathrm{d} \boldsymbol{x}=\left[-\frac{\beta_{t}}{2} \boldsymbol{x}-\beta_{t} \nabla_{\boldsymbol{x}_{t}} \log p_{t}\left(\boldsymbol{x}_{t}\right)\right] \mathrm{d} t+\sqrt{\beta_{t}} \mathrm{~d} \overline{\boldsymbol{w}} ,
\vspace{-0.1in}
\end{split}
\label{equ: sde}
\vspace{-0.1in}
\end{equation}
where $\beta_{t} \in (0,1)$ is the noise schedule, $t \in [0, T]$ represents the time index, and $\boldsymbol{w}$ and $\overline{\boldsymbol{w}}$ are the forward and backward Wiener processes, respectively. Once a DM is trained on large-scale clean data, diverse samples can be generated by replacing $\nabla_{\boldsymbol{x}_t} \log p_t(\boldsymbol{x}_t)$ in Eq.~\ref{equ: sde} with the learned score function $\boldsymbol{s}_\theta(\boldsymbol{x}_t, \sigma(t))$, where $\sigma(t)$ is the pre-defined noise strength corresponding to $\beta_t$.


By Bayes' rule, $p(\boldsymbol{x}|\boldsymbol{y}) \propto p(\boldsymbol{y}|\boldsymbol{x}) p(\boldsymbol{x})$, the reverse-time SDE can also be adapted to sample from the posterior distribution of images $\boldsymbol{x}$ conditioned on measurements $\boldsymbol{y}$. In this case, $p(\boldsymbol{y}|\boldsymbol{x})$ acts as a data-fidelity term to ensure the reconstructed images match the measurements. The score function from Eq.~\ref{equ: sde} is then modified as a conditional version:
\begin{equation}
\nabla_{\boldsymbol{x}_{t}} \log p_{t}\left(\boldsymbol{x}_{t}|\boldsymbol{y}\right)=\nabla_{\boldsymbol{x}_{t}} \log p_{t}\left(\boldsymbol{x}_{t}\right)+\nabla_{\boldsymbol{x}_{t}} \log p_{t}\left(\boldsymbol{y}|\boldsymbol{x}_{t}\right).
\label{equ:baye}
\end{equation}
where $p_{t}\left(\boldsymbol{y}|\boldsymbol{x}_{t}\right)$ is a time-dependent data-fidelity term distinct from $p(\boldsymbol{y}|\boldsymbol{x})$ and is generally intractable. The challenge in posterior sampling via diffusion models is bridging $p_{t}\left(\boldsymbol{y}|\boldsymbol{x}_{t}\right)$ with $p(\boldsymbol{y}|\boldsymbol{x})$. In the widely used diffusion posterior sampling (DPS) algorithm~\cite{chung2022diffusion}, we empirically assume that:
\begin{equation} 
\nabla_{\boldsymbol{x}_{t}} \log p_{t}\left(\boldsymbol{y}|\boldsymbol{x}_{t}\right) \approx \nabla{\boldsymbol{x}_{t}} \log p_{t}\left(\boldsymbol{y}|\hat{\boldsymbol{x}}_{0}\right), \text{where} \ \hat{\boldsymbol{x}}_{0} = \mathbb{E}[\boldsymbol{x}_{0}|\boldsymbol{x}_{t}]. 
\end{equation}
However, this approximation lacks theoretical guarantees for convergence to the true posterior distribution, which can result in inaccurate sampling, even in simple cases.


\subsection{Principled Expectation-Maximum Framework}

In this paper, we propose a principled expectation-maximization (EM) framework for learning clean diffusion models from noisy measurements. This approach inherits basic procedure of the standard EM algorithm~\cite{dempster1977maximum, gao2021deepgem} and EMDiffusion~\cite{bai2024expectation}, alternating between two steps: 1) the E-step, where we sample the posterior distribution of clean images from noisy measurements using a known diffusion model, and 2) the M-step, where we refine the DM using these recovered samples. This iterative process guides the DM towards a local minimum, progressively improving both the model and the quality of the recovered images.

\begin{algorithm}[tbp]
\begin{algorithmic}[1]
 \caption{Principled Expectation-Maximum Framework}
 \label{algo:algorithm1}
            \REQUIRE DM $\boldsymbol{s}_\theta$, measurements $(\boldsymbol{y}, f)$, few high-quality data $\boldsymbol{x}$, Iterations $N$, Timesteps $T$, schedule $\left\{\beta_t\right\}_{t=1}^T$
            \STATE Initialize $\boldsymbol{s}_\theta$ on $\boldsymbol{x}$ through DSM~\cite{vincent2011connection}
            \FOR{$i=1$ to $N$}
            \STATE $ \hat{\boldsymbol{x}}_0 \gets  \text{SamplingStep}(\boldsymbol{s}_\theta, \boldsymbol{y}, f, \left\{\beta_t\right\}_{t=1}^T)$ \textcolor{blue}{\COMMENT{Sec.~\ref{sec:3.2}}}
            \STATE $ \boldsymbol{s}_\theta \gets  \text{RefiningStep}(\boldsymbol{s}_\theta, \hat{\boldsymbol{x}}_0, \left\{\beta_t\right\}_{t=1}^T)$ \textcolor{blue}{\COMMENT{Sec.~\ref{sec:3.3}}}
            \ENDFOR
            \RETURN Learned $\boldsymbol{s}_\theta$
\vspace{-0.01in}
\end{algorithmic}
\end{algorithm}
\vspace{-0.02in}

\begin{table*}[htbp]  
  \centering  
  \caption{Quantitative comparisons of inverse imaging and learned priors. Best and second best results are highlighted in \textbf{bold} and \underline{underline}. DPS w/ clean prior is the upper bound of the results, as results in \colorbox{mygray} {gray} are inaccessible to the clean priors.}  
    \begin{tabular}{cccccccccc}  
    \toprule  
    \multirow{2}[4]{*}{\textbf{Method}} & \multicolumn{3}{c}{\textbf{CIFAR10-Denoising}} & \multicolumn{3}{c}{\textbf{CIFAR10-Inpainting}} & \multicolumn{3}{c}{\textbf{CelebA-Deblurring}} \\  
\cmidrule{2-10}          & PSNR$\uparrow $  & LPIPS$\downarrow $ & FID$\downarrow $   & PSNR$\uparrow $  & LPIPS$\downarrow $ & FID$\downarrow $   & PSNR$\uparrow $  & LPIPS$\downarrow $ & FID$\downarrow $ \\  
    \midrule  
    measurements & 18.05 & 0.047 & 132.59 & 13.49 & 0.295 & 234.47 & 22.47 & 0.365 & 72.83 \\  
   DPS w/ clean prior & 25.91 & 0.010 & 7.08 & 25.44 & 0.008 & 7.08 & 29.05 & 0.013 & 10.24 \\  
  \rowcolor{gray!20}  Noise2Self~\cite{batson2019noise2self} & 21.32 & 0.227 & {92.06} & - & - & - & - & - & - \\  
  \rowcolor{gray!20}  SURE-Score~\cite{aali2023solving}  & {22.42} & 0.138 & 132.61 & 15.75 & 0.182 & 220.01 & {22.07} & 0.383 & 191.96 \\  
  \rowcolor{gray!20}  Ambient Diffusion~\cite{daras2023ambient} & - & - & - & {20.57} & {0.027} & {28.88} & - & - & - \\  
 \rowcolor{gray!20}   EMDiffusion~\cite{bai2024expectation}  & \underline{23.16} & \underline{0.022} & \underline{86.47} & \textbf{24.70} & \textbf{0.009} & \textbf{21.08} & \underline{23.74} & \underline{0.103} & \underline{91.89} \\  
 \rowcolor{gray!20}   Ours  & \textbf{24.77} & \textbf{0.016} & \textbf{72.84} & \underline{22.74} & \underline{0.023} & \underline{25.46} & \textbf{26.58} & \textbf{0.081} & \textbf{79.43} \\  
    \bottomrule  
    \end{tabular}  
  \label{tab:results}%
  \vspace{-0.1in}
\end{table*}%

\begin{figure*}
	\centering
	\setlength{\tabcolsep}{1pt}
	\setlength{\fboxrule}{1pt}
        \resizebox{0.9\textwidth}{!}{
	\begin{tabular}{c}
		\begin{tabular}{cccccc|cccccc}
			\tiny{\makecell[c]{Noisy\\Measurement}} & 
			\tiny{\makecell[c]{SURE-\\Score~\cite{aali2023solving}}} & 
			\tiny{\makecell[c]{EM\\Diffusion~\cite{bai2024expectation}}} &
			\tiny{\makecell[c]{Ours}}&
			\tiny{\makecell[c]{DPS~\cite{chung2022diffusion} w/\\Clean Prior}}&
			\tiny{\makecell[c]{Ground\\Truth}} &
                \tiny{\makecell[c]{Masked\\Measurement}} & 
			\tiny{\makecell[c]{Ambient\\Diffusion~\cite{daras2023ambient}}} & 
			\tiny{\makecell[c]{EM\\Diffusion~\cite{bai2024expectation}}} &
			\tiny{\makecell[c]{Ours}}&
			\tiny{\makecell[c]{DPS~\cite{chung2022diffusion} w/\\Clean Prior}}&
			\tiny{\makecell[c]{Ground\\Truth}}
			\\ 
			\multicolumn{1}{c}{
				\begin{overpic}[width=0.076\linewidth]{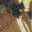}
				\end{overpic}
			}  &
                \multicolumn{1}{c}{
				\begin{overpic}[width=0.076\linewidth]{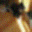}
				\end{overpic}
			}  &
                \multicolumn{1}{c}{
				\begin{overpic}[width=0.076\linewidth]{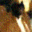}
				\end{overpic}
			}  &
			\multicolumn{1}{c}{
				\begin{overpic}[width=0.076\linewidth]{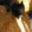}
				\end{overpic}
			}  &
			\multicolumn{1}{c}{
				\begin{overpic}[width=0.076\linewidth]{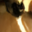}
				\end{overpic}
			}  &
			\multicolumn{1}{c|}{
				\begin{overpic}[width=0.076\linewidth]{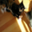}
				\end{overpic}
			}  &
			\multicolumn{1}{c}{
				\begin{overpic}[width=0.076\linewidth]{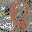}
				\end{overpic}
			}  &
			\multicolumn{1}{c}{
				\begin{overpic}[width=0.076\linewidth]{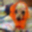}
				\end{overpic}
			}  &
			\multicolumn{1}{c}{
				\begin{overpic}[width=0.076\linewidth]{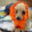}
				\end{overpic}
			}  &
                \multicolumn{1}{c}{
				\begin{overpic}[width=0.076\linewidth]{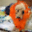}
				\end{overpic}
			}  &
                \multicolumn{1}{c}{
				\begin{overpic}[width=0.076\linewidth]{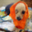}
				\end{overpic}
			}  &
			\multicolumn{1}{c}{
				\begin{overpic}[width=0.076\linewidth]{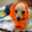}
				\end{overpic}
			} 
			\\
			  \multicolumn{1}{c}{
			  	\begin{overpic}[width=0.076\linewidth]{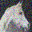}
			  	\end{overpic}
			  }  &
                \multicolumn{1}{c}{
				\begin{overpic}[width=0.076\linewidth]{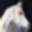}
				\end{overpic}
			}  &
                \multicolumn{1}{c}{
				\begin{overpic}[width=0.076\linewidth]{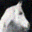}
				\end{overpic}
			}  &
			  \multicolumn{1}{c}{
			  	\begin{overpic}[width=0.076\linewidth]{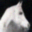}
			  	\end{overpic}
			  }  &
			  \multicolumn{1}{c}{
			  	\begin{overpic}[width=0.076\linewidth]{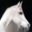}
			  	\end{overpic}
			  }  &
			  \multicolumn{1}{c|}{
			  	\begin{overpic}[width=0.076\linewidth]{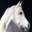}
			  	\end{overpic}
			  }  &
			  \multicolumn{1}{c}{
			  	\begin{overpic}[width=0.076\linewidth]{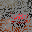}
			  	\end{overpic}
			  }  &
			  \multicolumn{1}{c}{
			  	\begin{overpic}[width=0.076\linewidth]{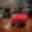}
			  	\end{overpic}
			  }  &
			  \multicolumn{1}{c}{
			  	\begin{overpic}[width=0.076\linewidth]{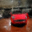}
			  	\end{overpic}
			  }  &
                \multicolumn{1}{c}{
			  	\begin{overpic}[width=0.076\linewidth]{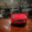}
			  	\end{overpic}
			  }  &
                \multicolumn{1}{c}{
			  	\begin{overpic}[width=0.076\linewidth]{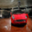}
			  	\end{overpic}
			  }  &
			  \multicolumn{1}{c}{
			  	\begin{overpic}[width=0.076\linewidth]{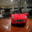}
			  	\end{overpic}
			  } 
			  \\
			  \multicolumn{1}{c}{
			  	\begin{overpic}[width=0.076\linewidth]{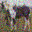}
			  	\end{overpic}
			  }  &
                \multicolumn{1}{c}{
				\begin{overpic}[width=0.076\linewidth]{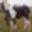}
				\end{overpic}
			}  &
                \multicolumn{1}{c}{
				\begin{overpic}[width=0.076\linewidth]{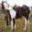}
				\end{overpic}
			}  &
			  \multicolumn{1}{c}{
			  	\begin{overpic}[width=0.076\linewidth]{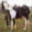}
			  	\end{overpic}
			  }  &
			  \multicolumn{1}{c}{
			  	\begin{overpic}[width=0.076\linewidth]{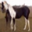}
			  	\end{overpic}
			  }  &
			  \multicolumn{1}{c|}{
			  	\begin{overpic}[width=0.076\linewidth]{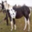}
			  	\end{overpic}
			  }  &
			  \multicolumn{1}{c}{
			  	\begin{overpic}[width=0.076\linewidth]{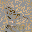}
			  	\end{overpic}
			  }  &
			  \multicolumn{1}{c}{
			  	\begin{overpic}[width=0.076\linewidth]{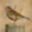}
			  	\end{overpic}
			  }  &
			  \multicolumn{1}{c}{
			  	\begin{overpic}[width=0.076\linewidth]{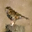}
			  	\end{overpic}
			  }  &
			  \multicolumn{1}{c}{
			  	\begin{overpic}[width=0.076\linewidth]{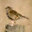}
			  	\end{overpic}
			  } &
                \multicolumn{1}{c}{
			  	\begin{overpic}[width=0.076\linewidth]{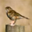}
			  	\end{overpic}
			  } &
                \multicolumn{1}{c}{
			  	\begin{overpic}[width=0.076\linewidth]{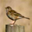}
			  	\end{overpic}
			  } \\
                \multicolumn{6}{c|}{(a) CIFAR10, Denoising} &
                \multicolumn{6}{c}{(b) CIFAR10, Inpainting} 
		\end{tabular}
	\end{tabular}}
	\caption{Qualitive results of image reconstruction on (a) CIFAR-10 noisy measurements and (b) CIFAR-10 masked measurements.}
\label{fig:exp-cifar-denoisinginpainting}
\vspace{-0.2in}
\end{figure*}

\subsubsection{E-step, principled DM-based posterior sampling}
\label{sec:3.2}
The E-step employs the plug-and-play Monte Carlo (PMC) algorithm for principled diffusion posterior sampling. Unlike conventional DPS, PMC is grounded in the theories of plug-and-play (PnP) optimization and Langevin Monte Carlo methods, providing non-asymptotic stationarity guarantees in terms of Fisher information. Assuming a minimum mean square error (MMSE) denoiser, $D_\sigma$, as an implicit image prior, the PnP framework is expressed as:
\vspace{-0.08in}
\begin{equation}
    \boldsymbol{x}_{t+1} = D_\sigma(\boldsymbol{x}_t + \gamma \nabla_{\boldsymbol{x_t}} \log p(\boldsymbol{y}|\boldsymbol{x}_t)),
\label{equ: pnp-denoiser}
\vspace{-0.08in}
\end{equation}
where $\gamma > 0$ is the step size, and $\sigma$ controls the denoising strength. According to Tweedie's formula~\cite{efron2011tweedie}, the MMSE denoiser can be formulated as:
\begin{equation}
\begin{split}
D_\sigma(\boldsymbol{x}) &= \boldsymbol{x} + \sigma^2 \nabla_{\boldsymbol{x}}\log p_\sigma(\boldsymbol{x}),\\
\text{where} \ p_\sigma(\boldsymbol{x}) &= \int p(\boldsymbol{\mu}) \mathcal{N}(\boldsymbol{x}-\boldsymbol{\mu}; 0, \sigma \mathbf{I}) d \boldsymbol{\mu}.
\end{split}
\label{equ: tweedie}
\end{equation}
Here, $p_\sigma(\boldsymbol{x})$ represents the prior distribution convolved with Gaussian noise of standard deviation $\sigma$. Though intractable, this prior can be approximated by the score-based diffusion model $\boldsymbol{s}_\theta(\boldsymbol{x}, \sigma)$. Consequently, the PnP optimization in Eq.~\ref{equ: pnp-denoiser} can be rewritten as:
\vspace{-0.08in}
\begin{equation}
\begin{split}
    \boldsymbol{x}_{t+1} =& \boldsymbol{x}_t + \gamma \nabla_{\boldsymbol{x_t}} \log p(\boldsymbol{y}|\boldsymbol{x}_t) \\
    & + \sigma^2 \nabla_{\boldsymbol{x}}\log p_\sigma(\boldsymbol{x}_t + \gamma \nabla_{\boldsymbol{x_t}} \log p(\boldsymbol{y}|\boldsymbol{x}_t)),\\
    =& \boldsymbol{x}_t + \gamma \nabla_{\boldsymbol{x_t}} \log p(\boldsymbol{y}|\boldsymbol{x}_t) \\
    & - \sigma^2 \boldsymbol{s}_\theta(\boldsymbol{x}_t+\gamma \nabla_{\boldsymbol{x}_t} \log p(\boldsymbol{y}|\boldsymbol{x}_t), \sigma(t)).
\end{split}
\label{equ: pnp-denoiser2}
\vspace{-0.08in}
\end{equation}
Introducing additional Brownian motion $\boldsymbol{\omega}_t$ and a scaling factor $\alpha_k$ before the score function leads to the PMC formulation, as originally described in \cite{sun2023provable}:
\begin{equation}
\begin{split}
    \boldsymbol{x}_{t+1} =& \boldsymbol{x}_t + \gamma \nabla_{\boldsymbol{x_t}} \log p(\boldsymbol{y}|\boldsymbol{x}_t) \\
    & - \alpha_k \boldsymbol{s}_\theta(\boldsymbol{x}_t+\gamma \nabla_{\boldsymbol{x}_t} \log p(\boldsymbol{y}|\boldsymbol{x}_t), \sigma(t)) + \boldsymbol{\omega}_t,
\end{split}
\label{equ: pmc}
\vspace{-0.08in}
\end{equation}
where $k$ index the EM iterations. As the diffusion model's accuracy improves, $\alpha_k$ increases from a small value (e.g., $1e-3$) to 1. Since PMC does not rely on approximations for the time-dependent data fidelity $p_{t}\left(\boldsymbol{y}|\boldsymbol{x}_{t}\right)$, it ensures principled convergence to the true posterior distribution.

\subsubsection{M-step, refining diffusion priors}
\label{sec:3.3}
After obtaining the posterior samples of clean images (Sec.~\ref{sec:3.2}), we refine the diffusion model using these reconstructed images. The M-step is similar to standard diffusion model training with clean data and uses the denoising score matching (DSM) loss to update the model parameters:
\vspace{-0.15in}
\begin{equation}
    \theta^{*}=\underset{\theta}{\arg\min } \mathbb{E}_{t, \boldsymbol{x}_t, \hat{\boldsymbol{x}}}\left[\left\|\boldsymbol{s}_{\theta}(\boldsymbol{x}_t, t)-\nabla_{\boldsymbol{x}_{t}} \log p(\boldsymbol{x}_t | \hat{\boldsymbol{x}})\right\|_{2}^{2}\right],
\label{equ:obj-ddpm}
\end{equation}
where $\hat{\boldsymbol{x}}$ represents the posterior samples, and $\boldsymbol{x}_t$ are the noisy versions of $\hat{\boldsymbol{x}}_0$ generated by the forward SDE in Eq.~\ref{equ: sde}.

To speed up training, we do not retrain the diffusion model from scratch in each EM iteration. Instead, we fine-tune the model parameters $\theta$ from the previous iteration during the early stages. In the final 1-3 EM iterations, we reset the model to eliminate any influence of the low-quality posterior samples from earlier iterations, then retrain it from scratch.



\subsubsection{Initialization of the EM iterations}
A good initialization of DMs is crucial for ensuring the EM algorithm converges to the correct local minimum. In our method, we pre-train the DMs using 50 randomly selected clean images before the first EM iteration.


\section{Experiments}

\subsection{Experimental Settings}

\textbf{Datasets.}
We evaluate our method on two datasets with distinct characteristics: CIFAR10~\cite{krizhevsky2009learning} at a resolution of $32\times32$ and CelebA~\cite{liu2015faceattributes} at $64\times64$. Noisy measurements $\boldsymbol{y}$ are generated by corrupting clean images $\boldsymbol{x}$ from the training sets. 
In the first six iterations, 5,000 measurements are randomly selected for posterior sampling and diffusion model refinement.
In the last three iterations, all measurements are used for posterior sampling and training diffusion model from scratch.
We randomly choose 250 clean images from test sets for evaluation of image reconstruction and original training sets for evaluation of image generation.

\textbf{Baselines.}
We compare our method with four related baselines: Ambient Diffusion~\cite{daras2023ambient}, SURE-Score~\cite{aali2023solving}, EMDiffusion~\cite{bai2024expectation}, and DPS with a clean prior~\cite{chung2022diffusion}. Ambient Diffusion, SURE-Score, and EMDiffusion have similar settings to our method, as they do not rely on large-scale clean data to train DMs, but instead train DMs directly from noisy measurements. In contrast, DPS~\cite{chung2022diffusion} leverages a pre-trained clean diffusion prior for posterior sampling, representing the performance upper bound for our approach.

\begin{figure*}[htbp]
	\centering
	\setlength{\tabcolsep}{1pt}
	\setlength{\fboxrule}{1pt}
        \resizebox{0.74\textwidth}{!}{
	\begin{tabular}{c}
		\begin{tabular}{cccccccccc}
			\tiny{\makecell[c]{Masked\\Measurement}} & 
			\tiny{\makecell[c]{SURE-\\Score~\cite{aali2023solving}}} & 
			\tiny{\makecell[c]{EM\\Diffusion~\cite{bai2024expectation}}} &
			\tiny{\makecell[c]{Ours 1st\\Iteration}}& 
                \multicolumn{3}{c}{\begin{tikzpicture}[baseline]  
                    \draw[->, >=latex, line width=0.35mm] (0,0.1) -- (3.7cm,0.1);  
                    \end{tikzpicture}  } &
                \tiny{\makecell[c]{Ours Final\\Iteration}}&
			\tiny{\makecell[c]{DPS~\cite{chung2022diffusion} w/\\Clean Prior}}&
			\tiny{\makecell[c]{Ground\\Truth}}
			\\ 
			\multicolumn{1}{c}{
				\begin{overpic}[width=0.092\linewidth]{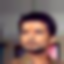}
				\end{overpic}
			}  &
                \multicolumn{1}{c}{
				\begin{overpic}[width=0.092\linewidth]{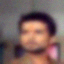}
				\end{overpic}
			}  &
                \multicolumn{1}{c}{
				\begin{overpic}[width=0.092\linewidth]{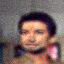}
				\end{overpic}
			}  &
			\multicolumn{1}{c}{
				\begin{overpic}[width=0.092\linewidth]{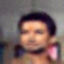}
				\end{overpic}
			}  &
			\multicolumn{1}{c}{
				\begin{overpic}[width=0.092\linewidth]{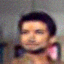}
				\end{overpic}
			}  &
			\multicolumn{1}{c}{
				\begin{overpic}[width=0.092\linewidth]{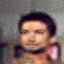}
				\end{overpic}
			}  &
			\multicolumn{1}{c}{
				\begin{overpic}[width=0.092\linewidth]{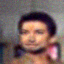}
				\end{overpic}
			}  &
			\multicolumn{1}{c}{
				\begin{overpic}[width=0.092\linewidth]{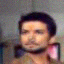}
				\end{overpic}
			}  &
			\multicolumn{1}{c}{
				\begin{overpic}[width=0.092\linewidth]{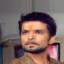}
				\end{overpic}
			}  &
			\multicolumn{1}{c}{
				\begin{overpic}[width=0.092\linewidth]{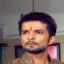}
				\end{overpic}
			} 
			\\
			  \multicolumn{1}{c}{
			  	\begin{overpic}[width=0.092\linewidth]{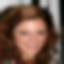}
			  	\end{overpic}
			  }  &
                \multicolumn{1}{c}{
				\begin{overpic}[width=0.092\linewidth]{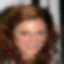}
				\end{overpic}
			}  &
                \multicolumn{1}{c}{
				\begin{overpic}[width=0.092\linewidth]{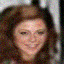}
				\end{overpic}
			}  &
			  \multicolumn{1}{c}{
			  	\begin{overpic}[width=0.092\linewidth]{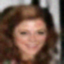}
			  	\end{overpic}
			  }  &
			  \multicolumn{1}{c}{
			  	\begin{overpic}[width=0.092\linewidth]{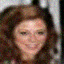}
			  	\end{overpic}
			  }  &
			  \multicolumn{1}{c}{
			  	\begin{overpic}[width=0.092\linewidth]{fig/fig3_emscore/2-6.png}
			  	\end{overpic}
			  }  &
			  \multicolumn{1}{c}{
			  	\begin{overpic}[width=0.092\linewidth]{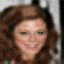}
			  	\end{overpic}
			  }  &
			  \multicolumn{1}{c}{
			  	\begin{overpic}[width=0.092\linewidth]{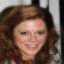}
			  	\end{overpic}
			  }  &
			  \multicolumn{1}{c}{
			  	\begin{overpic}[width=0.092\linewidth]{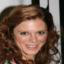}
			  	\end{overpic}
			  }  &
			  \multicolumn{1}{c}{
			  	\begin{overpic}[width=0.092\linewidth]{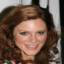}
			  	\end{overpic}
			  } 
			  \\
			  \multicolumn{1}{c}{
			  	\begin{overpic}[width=0.092\linewidth]{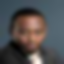}
			  	\end{overpic}
			  }  &
                \multicolumn{1}{c}{
				\begin{overpic}[width=0.092\linewidth]{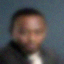}
				\end{overpic}
			}  &
                \multicolumn{1}{c}{
				\begin{overpic}[width=0.092\linewidth]{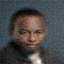}
				\end{overpic}
			}  &
			  \multicolumn{1}{c}{
			  	\begin{overpic}[width=0.092\linewidth]{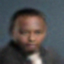}
			  	\end{overpic}
			  }  &
			  \multicolumn{1}{c}{
			  	\begin{overpic}[width=0.092\linewidth]{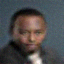}
			  	\end{overpic}
			  }  &
			  \multicolumn{1}{c}{
			  	\begin{overpic}[width=0.092\linewidth]{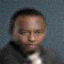}
			  	\end{overpic}
			  }  &
			  \multicolumn{1}{c}{
			  	\begin{overpic}[width=0.092\linewidth]{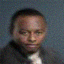}
			  	\end{overpic}
			  }  &
			  \multicolumn{1}{c}{
			  	\begin{overpic}[width=0.092\linewidth]{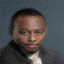}
			  	\end{overpic}
			  }  &
			  \multicolumn{1}{c}{
			  	\begin{overpic}[width=0.092\linewidth]{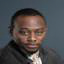}
			  	\end{overpic}
			  }  &
			  \multicolumn{1}{c}{
			  	\begin{overpic}[width=0.092\linewidth]{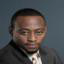}
			  	\end{overpic}
			  } 
		\end{tabular}
	\end{tabular}}
	\vspace{-0.1cm}
\caption{{Qualitive results on CelebA deblurring.} For each image, the blur kernel is a $9\times9$ Gaussian blur kernel. Within the principled iterative learning framework, the diffusion model learns cleaner score-based priors, improving the quality of reconstructed images.}
\label{exp:cifar-deblurring}
\vspace{-0.1in}
\end{figure*}
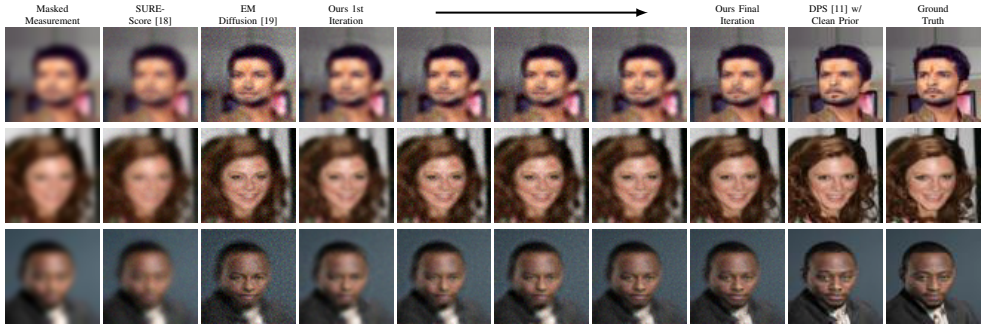

\begin{figure*}[htbp]
	\centering
	\setlength{\tabcolsep}{3pt}
	\setlength{\fboxrule}{1pt}
	\begin{tabular}{c}
		\begin{tabular}{ccc} 
                \multicolumn{1}{c}{
				\begin{overpic}[width=0.26\linewidth]{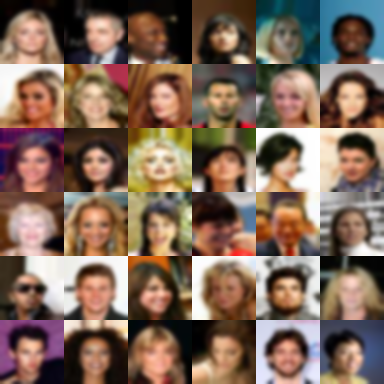}
				\end{overpic}
			}  &
			\multicolumn{1}{c}{
				\begin{overpic}[width=0.26\linewidth]{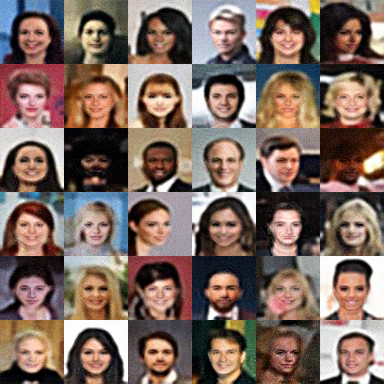}
				\end{overpic}
			}  &
			\multicolumn{1}{c}{
				\begin{overpic}[width=0.26\linewidth]{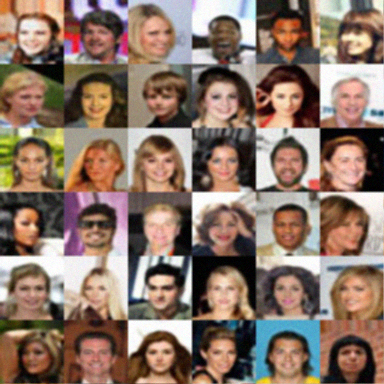}
				\end{overpic}
			} 
                \vspace*{-0.03cm}
                \\
                \multicolumn{1}{c}{Training Data} &
                \multicolumn{1}{c}{EMDiffusion} &
                \multicolumn{1}{c}{Ours} 
		\end{tabular}
        \end{tabular}
	\caption{
 {Comparison of uncurated samples generated from diffusion models trained by EMDiffusion and the proposed method.}}
	\label{fig:exp-generation}
	\vspace*{-0.22in}
\end{figure*}

\textbf{Metrics.}
We evaluate image reconstruction quality using two metrics: peak signal-to-noise ratio (PSNR) and learned perceptual image patch similarity (LPIPS). Additionally, we compute the Fréchet Inception Distance (FID) between samples generated by the trained DMs and the reserved clean training set to assess the quality of image generation.
\vspace*{-0.05in}

\subsection{Image Reconstruction}
We evaluate our principled expectation-maximization (EM) framework on two datasets across three representative imaging inverse problems: image denoising, random inpainting, and Gaussian deblurring.

\textbf{Image denoising.}
For image denoising, we add Gaussian noise $\boldsymbol{n} \sim \mathcal{N}(0, \sigma^2\boldsymbol{I})$ with $\sigma = 0.2$. As shown in Table~\ref{tab:results} and Fig.~\ref{fig:exp-cifar-denoisinginpainting}, our method significantly outperforms all baselines. The reconstructed images exhibit notably fewer artifacts compared to the current state-of-the-art, EMDiffusion, demonstrating the advantage of moving beyond empirical approximations for time-dependent data fidelity.


\textbf{Image inpainting.}
For image inpainting, we randomly mask 60\% of the pixels in the images. As shown in Table~\ref{tab:results} and Fig.~\ref{fig:exp-cifar-denoisinginpainting}, the posterior samples closely match the ground truths, highlighting the robustness of our method. Although our method achieves the second-best performance in the random inpainting task, this is due to EMDiffusion’s simple approximation being particularly effective for this problem~\cite{dou2024diffusion}. In EMDiffusion, the measurement $\boldsymbol{y}$ guides the updates of all pixels, through $\nabla_{\boldsymbol{x}_{t}} \log p_{t}\left(\boldsymbol{y}|\hat{\boldsymbol{x}}_{0}\right)$, not just the unmasked ones.

\textbf{Image deblurring.}
For image deblurring, we apply a $9 \times 9$ Gaussian blur kernel with a standard deviation of 2. As shown in Table.~\ref{tab:results} and Fig.~\ref{exp:cifar-deblurring}, our method consistently outperforms all baselines. The posterior samples progressively approximate the ground truths throughout the iterations, verifying the effectiveness of the EM framework.

\vspace*{-0.05in}
\subsection{Learned Diffusion Priors}
In Table~\ref{tab:results} and Fig.~\ref{fig:exp-generation}, we compare the quantitative and qualitative results of the diffusion priors learned by our method against the baselines. The FID scores in Table~\ref{tab:results} demonstrate that our method outperforms the baselines in all three tasks: image inpainting and denoising on CIFAR10, and image deblurring on CelebA. The progressively improving quality of the posterior samples in Fig.~\ref{exp:cifar-deblurring} suggests that the diffusion model is effectively approximating the true prior distribution, validating the effectiveness of our iterative learning framework. As shown in uncurated samples in Fig.~\ref{fig:exp-generation}, our method significantly reduces sharp artifacts compared to EMDiffusion, further confirming its superior performance. 

\vspace*{-0.05in}
\section{Conclusion}
In this paper, we present a principled expectation-maximization framework for learning clean diffusion models from partial measurements. Each iteration of the framework employs a plug-and-play Monte Carlo algorithm to reconstruct posterior image samples from the measurements, followed by refining the diffusion model with these recovered images. Our method demonstrates superior performance across various tasks and datasets, outperforming existing baselines in both image reconstruction and diffusion model training. For future work, we aim to extend this approach to more challenging scenarios, such as extremely sparse or noisy measurements in scientific applications.




\bibliographystyle{IEEEtran}
\bibliography{IEEEabrv,reference}

\end{document}